\documentclass[11pt]{article}

\usepackage[preprint]{acl}

\usepackage{times}
\usepackage{latexsym}

\usepackage[T1]{fontenc}

\usepackage[utf8]{inputenc}

\usepackage{microtype}

\usepackage{inconsolata}

\usepackage{graphicx}
\usepackage{booktabs}
\usepackage{enumitem}
\usepackage{amsmath}
\usepackage{subcaption}
\usepackage{xcolor}
\usepackage{cleveref}
\usepackage{fancyvrb}
\usepackage{stfloats}
\usepackage{wrapfig} 

\usepackage{amssymb}
\usepackage{fvextra}
\usepackage[table,dvipsnames]{xcolor}

\definecolor{headerbg}{RGB}{250,250,250}
\definecolor{oursbg}{RGB}{238,247,238}
\definecolor{oursbgstrong}{RGB}{229,242,229}
\definecolor{variantbg}{RGB}{247,247,247}

\definecolor{oursgreen}{RGB}{34,120,55}
\definecolor{sectiongray}{RGB}{90,90,90}

\title{Plan2Map: A Multimodal Benchmark for Document-Grounded Geospatial Boundary Reconstruction from Planning Records}

\author{
  \mdseries Fabian Degen\textsuperscript{1,*},
  Oishi Deb\textsuperscript{1,*},\\
  Jindong Gu\textsuperscript{1,2},
  Junchi Yu\textsuperscript{1}, 
  Samuele Marro\textsuperscript{1}, \\
  Philip Torr\textsuperscript{1},
  Jialin Yu\textsuperscript{1, $\dagger$}
  \\
  \textsuperscript{1}{University of Oxford}
  \textsuperscript{2}{Google DeepMind}
  \\
  \small{
  \href{mailto:fabian.degen@tuta.com}{fabian.degen@tuta.com}
  \quad
    \href{mailto:oishideb@robots.ox.ac.uk}{oishideb@robots.ox.ac.uk}
    \quad
    \href{mailto:yu.jialin@outlook.com}{yu.jialin@outlook.com}
  }
  \\
  \small{
    \textsuperscript{*}Equal Contribution.\quad
    \textsuperscript{$\dagger$}Corresponding authors
  }
}

\begin{document}
\maketitle   

\rowcolors{2}{gray!15}{white}      

\begin{abstract}
Planning records define restrictions over geographic areas, but their source documents often provide only indirect spatial evidence rather than machine-readable boundaries. We introduce \textbf{Plan2Map}, a 208-case multimodal benchmark for document-grounded geospatial boundary reconstruction from UK planning records. Given only a source planning document, systems must reconstruct a valid geospatial boundary from notice text, schedules, map plates, map labels, and boundary annotations; the reference GeoJSON is held out for scoring. We propose \textbf{GeoPlanAgent}, a document-grounded, geospatial-tool-in-the-loop system that decomposes the task into evidence extraction, localisation, map registration, boundary segmentation, projection, and verification. On Plan2Map, GeoPlanAgent achieves 0.736 mean IoU and 0.904 median IoU, with 67.8\% of predictions at or above 0.8 IoU, substantially outperforming direct VLM-to-GeoJSON baselines. Diagnostic analysis shows that direct VLM prediction remains unreliable, while remaining errors are concentrated in localisation and map registration, and supervised boundary segmentation substantially improves pixel-level mask quality. Plan2Map provides a concrete testbed for multimodal geospatial reconstruction from public planning records. Project page: \url{https://odeb1.github.io/Plan2Map_Project_Page/}
\end{abstract}

\section{Introduction}
\label{section:intro}

Planning records often answer a spatial question: \emph{which area is subject to a particular rule?} For the UK Article~4 Directions \citep{planningdata2026article4direction}, the affected area is typically described through a legal notice and an accompanying map rather than provided as machine-readable geometry. Digital planning systems, however, need such geometry to check whether a site or application falls inside an affected area, compare restrictions with other planning datasets, and audit or update records over time. The source documents usually provide only indirect spatial evidence: notice text, legal schedules, scanned or embedded map plates, map labels, coloured or hatched regions, and boundary annotations. Recovering the boundary is therefore not simply field extraction, but reconstruction of a structured spatial object from distributed textual, visual, and geographic evidence.

\begin{figure}[t]
    \centering
    \vspace{-5mm}
    \resizebox{0.47\textwidth}{!}{
    \includegraphics[]{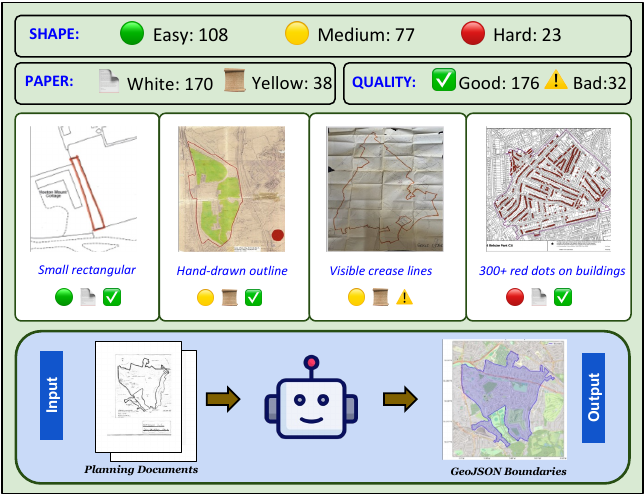}
    }
    \caption{\textbf{Plan2Map overview.} Plan2Map pairs source planning documents with verified GeoJSON boundaries and includes cases spanning different boundary shapes, document formats, and scan qualities. Given only the source document as input, a system must reconstruct the corresponding geospatial boundary.}
    \label{fig:teaser}
    \vspace{-5mm}
\end{figure}

This reconstruction problem is practically important for planning-record digitisation, where government and industry efforts such as the ``Extract initiative'' seek to convert legacy planning records into structured digital data \citep{govuk2025extract,colebourn2025extract}. Recent reviews of planning-application delays further underscore the need for efficient planning information flows \citep{weinstein2024UKgov,mhclg2026planningdelays}. Although public planning-data portals\footnote{https://www.planning.data.gov.uk/entity/?dataset=article-4-direction-area} may publish spatial records for some Article~4 Directions, including GeoJSON geometries, these records should be understood as evaluation references rather than task inputs: in our setting, the system is given only the source planning document. The challenge is to recover the boundary that a human would infer from the document itself, not to retrieve or copy a pre-existing geometry.


We frame this setting as \emph{document-grounded geospatial boundary reconstruction} and introduce \textbf{Plan2Map}, a multimodal benchmark for reconstructing machine-readable geospatial boundaries from planning documents. As demonstrated in Fig.\ref{fig:teaser}, each example pairs a UK Article~4 Direction source document with a verified reference geometry. At evaluation time, the only case-specific input is the source planning document: the reference GeoJSON is held out for scoring, while systems may use declared public geospatial resources such as gazetteers and basemaps. Unlike standard document-understanding tasks that return textual answers, tables, or labels, Plan2Map requires a system to transform dispersed textual and visual evidence into an externally verifiable geospatial object.

This task is challenging because the boundary is not stated in machine-readable coordinates. A system must move from planning-document evidence to geographic grounding and finally to polygon reconstruction, while handling heterogeneous PDFs containing scanned pages, embedded maps, ambiguous boundary symbols, and textual qualifications. These properties make Plan2Map a stress test for multimodal systems that must produce structured spatial outputs rather than textual responses alone. To establish a strong baseline, we propose \textbf{GeoPlanAgent}, a decomposition-based reference system for Plan2Map. GeoPlanAgent separates the problem into document reading, localization, map registration, boundary extraction, projection, and verification, allowing us to diagnose which stages limit current multimodal systems.
 
Our \textbf{main contributions} are: \textbf{(i)} We introduce \textbf{Plan2Map}, a 208-case, manually reviewed benchmark for document-grounded geospatial boundary reconstruction, built from UK Article~4 source documents and verified boundary geometries; \textbf{(ii)} We define the associated task and evaluation protocol: given only the source planning document, a system must reconstruct a valid geospatial boundary, evaluated by geometric overlap, centroid error, and cost; \textbf{(iii)} We propose \textbf{GeoPlanAgent}, a decomposition-based system that mirrors the task structure through reading, localization, map registration, boundary segmentation, projection, and verification; and \textbf{(iv)} We provide diagnostic analysis with direct VLM baselines and component ablations, showing that localization and boundary extraction are the dominant bottlenecks.


\section{Related Work}

\paragraph{Multimodal Document and Geospatial Reasoning.}

Recent work has made substantial progress on multimodal document understanding and geospatial reasoning, but these capabilities have mostly been evaluated in separate settings. Document-understanding benchmarks such as DocVQA test whether models can read document images and answer questions over text, layout, and visual structure, while LayoutLMv3 and Donut represent layout-aware and OCR-free approaches to document understanding \citep{mathew2021docvqa,huang2022layoutlmv3,kim2022donut}. Geospatial VQA and remote-sensing benchmarks such as RSVQA, EarthVQA, GeoChat, and EarthWhere evaluate visual question answering, grounded dialogue, and localization over maps, satellite imagery, or street-level images \citep{lobry2020rsvqa,wang2024earthvqa,kuckreja2024geochat,qian2025earthwhere}. Related map-digitisation and georeferencing literature, including MapReader and historical-map vectorisation benchmarks, studies scanned-map analysis, patch classification, text spotting, vectorisation, and control-point alignment \citep{wood2024mapreader,chen2024historicalmapvectorization}. In contrast, Plan2Map requires a system to read a source planning PDF, identify spatial evidence across text and maps, ground the relevant map in real geography, and output a valid geospatial boundary.


\paragraph{Planning-record digitisation.}
Government and industry efforts such as the UK Extract initiative highlight the practical need to convert existing planning records into structured data for digital planning tools \citep{govuk2025extract,colebourn2025extract,google2025extract}. These efforts motivate Plan2Map's setting, but they are not released as reusable academic benchmarks with source-document inputs, reference geometries, output schemas, and scoring protocols. Urban-planning and legal benchmarks evaluate policy knowledge or textual reasoning \citep{zheng2025urbanplanbench,chalkidis2022lexglue,guha2023legalbench}, but they do not require systems to recover a geospatial boundary from planning-document evidence.

\paragraph{Tool-augmented and decomposition-based systems.}
Plan2Map also relates to tool-augmented language-model systems such as retrieval-augmented generation, ReAct, and Toolformer \citep{lewis2020rag,yao2023react,schick2023toolformer}. GeoPlanAgent adapts this idea to geospatial reconstruction: rather than asking a VLM to produce a polygon in one pass, it decomposes the task into document-evidence extraction, localization with geocoders, map registration, boundary segmentation, projection, and verification. Our experiments show that this decomposition is necessary: direct VLM PDF-to-GeoJSON prediction performs poorly even with frontier models.

\section{Dataset}
\label{sec:dataset}

We curate \textbf{Plan2Map}, a benchmark of 208 UK Article~4 Direction records. Article~4 Directions are planning instruments through which local planning authorities withdraw permitted-development rights for specified buildings or areas \citep{mhclg2024planningpracticeguidance}. We source records from open UK planning datasets released under the Open Government Licence. Each case pairs a source planning PDF with a verified reference boundary in GeoJSON and a rendered location-map visualisation for human inspection. The dataset spans 1958--2025 and covers 29 local planning authorities across England.

\paragraph{Sources and Construction}

Each raw entry contains an Article~4 Direction document and a corresponding spatial boundary from the public planning record. We manually reviewed these pairs to ensure that the boundary represented in the PDF matched the supplied GeoJSON. The curation pipeline reduced 270 raw entries to 208 release-ready cases: we removed 40 records where the PDF boundary and supplied GeoJSON disagreed, 9 records whose PDFs contained no usable boundary map, and 7 duplicates; we also consolidated 5 multi-GeoJSON groups where the same PDF was distributed across several authority entries. All released cases therefore have a source document whose spatial evidence can be visually matched to the held-out reference geometry.

\paragraph{Dataset Composition}

Each released item contains three artefacts: the \textit{PDF}, which provides the planning-document text, schedules, maps, labels, and boundary annotations; the \textit{GeoJSON}, a \texttt{Polygon} or \texttt{MultiPolygon} used only as the evaluation reference; and a \textit{location-map PNG}, which overlays the reference geometry on an OpenStreetMap basemap for inspection. We also provide metadata describing the local authority, site description, document quality, boundary shape, shape complexity, and validation status.

\paragraph{Distribution}

Plan2Map includes scanned and born-digital documents, varied map quality, and boundaries ranging from simple parcels to irregular or multi-part geometries. These properties make it a compact but non-trivial benchmark for document-grounded geospatial boundary reconstruction. Full metadata definitions, geographic distributions, document-quality labels, shape taxonomy, and curation details are provided in Appendix~\ref{appendix:dataset-details}.

\section{GeoPlanAgent}
\label{section:geoplanagent}

\begin{figure*}[t]
    \centering
    \vspace{-15mm}
    \resizebox{0.99\textwidth}{!}{
    \includegraphics[]{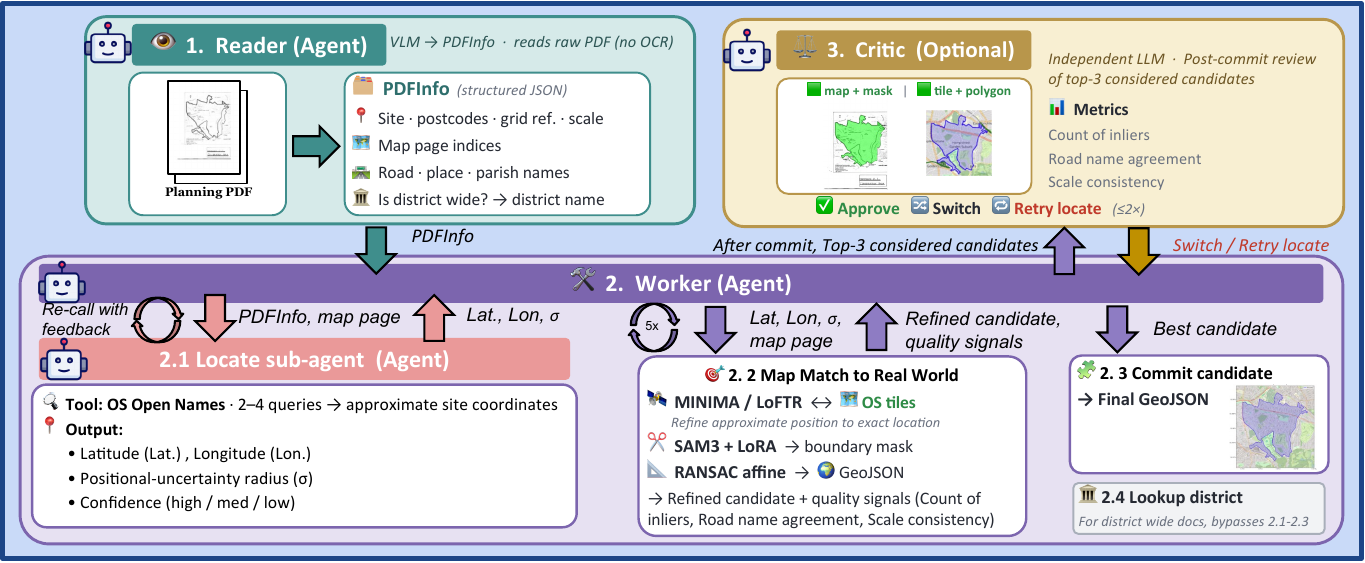}
    }
\caption{\textbf{GeoPlanAgent workflow.}
GeoPlanAgent decomposes Plan2Map into document evidence extraction, geographic localisation, map-to-basemap registration, boundary segmentation, and GeoJSON projection. The Reader converts the source PDF into structured spatial evidence; the Worker calls the Locate sub-agent, runs map matching and boundary extraction tools, and commits a candidate GeoJSON. The optional Critic reviews candidate matches and may approve, switch, or request re-localisation.}
    \label{fig:agent_overview}
    \vspace{-5mm}
\end{figure*}

In this section, we present \textbf{GeoPlanAgent}, a tool-augmented agentic system for the \textbf{Plan2Map} task. Rather than asking a multimodal model to generate a geospatial polygon in one pass, GeoPlanAgent decomposes the task into document evidence extraction, geographic localisation, and map-to-boundary reconstruction. The system consists of two primary agents, a Locate sub-agent, and an optional Critic, as illustrated in \Cref{fig:agent_overview}. The \emph{Reader} converts the source planning document into a typed record of spatial evidence. The \emph{Worker} uses this record to call the Locate sub-agent, register the planning map against a national basemap, segment the drawn boundary, project the result into geospatial coordinates, and return a valid GeoJSON geometry. The optional \emph{Critic} performs a second pass of visual verification over candidate matches.

All agents use Gemini~3~Flash by default~\cite{google2025geminiflash3} at temperature~0 for reproducibility. They are implemented with \texttt{pydantic-ai}, the agent framework built on Pydantic~\cite{Colvin_Pydantic_Validation_2026}, and each agent's final output is parsed against a Pydantic schema before being passed downstream.


\subsection{Reader}
\label{subsection:reader}

The Reader is a single multimodal LLM call over the raw PDF. It extracts spatial signals needed downstream, including postcodes, OS grid references, site addresses, road and place names, administrative regions, visible map labels, printed map scale, and per-page metadata indicating which pages contain positionable maps. The Reader also sets a Boolean flag that indicates whether the planning PDF covers an entire district. This flag is used to bypass the pipeline described below and directly look up the district boundary in geospatial coordinates, which improves both speed and performance. Full schema fields and prompts are provided in \Cref{appendix:text-extraction,appendix:prompts}.


\subsection{Worker}
\label{subsection:worker}

The Worker produces the final GeoJSON boundary via a tool-calling loop. It receives the structured PDF information extracted from the Reader and the map page. For the subsequent steps to work well, the map needs to be north-aligned. To do this, we fine-tuned a ResNet50~\cite{ResNet50} rotation classifier (\Cref{subsection:rotation-classifier}), which corrects rotated maps before the loop begins. The Worker operates over four tools:

\paragraph{Tool 1: Approximate map-center.}

The first tool allows the Worker to invoke the Locate sub-agent to obtain coordinates for a candidate map centre. The Locate sub-agent sees the structured PDF information extracted by the Reader and a rendered image of the map page displaying the planning site. The Locate sub-agent then returns the approximate coordinates of the center of the map displaying the planning site, an uncertainty radius $\sigma$, a confidence label, and a short evidence string. Its only tool is an offline OS Open Names~\cite{os_open_names} gazetteer query; the sub-agent composes 2--4 queries by combining signals extracted from the Reader (site address, parish, place names, administrative region) with labels it reads off the map image directly, then clusters the returned candidates and picks one. The radius $\sigma$ tracks confidence: $\approx 200$\,m when multiple queries agree within 500\,m, 800--1500\,m for a single ambiguous pick. 

The Worker may re-invoke the sub-agent with a feedback string after a weak match; the sub-agent receives that feedback alongside its prior reasoning and is prompted to pick a different signal type (e.g., a map-visible landmark rather than a postcode that could be from the District Council rather than the planning site). Additional geocoder variants are described in \Cref{appendix:locate-tools}.

\paragraph{Tool 2: Map-tile matching.}

The Locate sub-agent provides only an approximate centre: even a confident pick has uncertainty on the order of hundreds of metres, while many planning boundaries are parcel-sized. The \texttt{match} tool therefore refines this coarse location by aligning the rendered planning map against OS Open Zoomstack \cite{os_open_zoomstack} basemap tiles and then segmenting and projecting the boundary through the recovered alignment. The Worker can re-invoke this tool a maximum number of five times for different candidate centres. Each \texttt{match} call returns a single candidate, so across the five-call budget the Worker accumulates up to five candidates before committing the best one. A single \texttt{match} call performs three steps.

\textbf{(1) Sliding-window search.} Given a candidate centre and uncertainty radius $\sigma$, \texttt{match} renders a tile canvas around the candidate location and searches for the best alignment between the planning map and the basemap. The search window expands with $\sigma$: tight Locate predictions require only a small local search, while ambiguous predictions search a larger area. To make visual matching scale-consistent, the planning map is resized using the printed map scale when available; otherwise, the search sweeps common UK planning-map scales. We use MINIMA-LoFTR~\cite{Sun_2021_CVPR,Ren_2025_CVPR} to match features between the planning map and OS Open Zoomstack tiles, fit candidate affine transforms with RANSAC~\cite{Ransac}, and rank alignments by inlier strength and spatial consistency. When the Reader extracts road names, we additionally favour alignments whose nearby OS road graph contains those names. The highest-scoring alignment is retained. Full scale-sweep and re-ranking details are in \Cref{subsection:matching}.

\textbf{(2) Boundary segmentation.} After the sliding-window search, the boundary mask on the map image is segmented by SAM~3~\cite{carion2025sam3segmentconcepts} fine-tuned on human-annotated planning maps with LoRA~\cite{hu2022lora} adapters --- a parameter-efficient method that trains small low-rank weight updates on top of the frozen SAM~3 backbone. To widen the visual diversity beyond the small annotated pool, we employ style-transfer data augmentations, as described in \Cref{subsection:style-transfer}. Because the training pool overlaps with the benchmark cases, the fine-tune is run as 5-fold cross-validation: each evaluation case is segmented by the adapter from the fold that excluded it, so no case is ever scored by a model that has seen its ground truth. The contribution of SAM3-LoRA over the base SAM3 and segmentation by the LLM directly is quantified in \Cref{subsection:abl-finetune}, and training details are in \Cref{subsection:sam3-finetune}.

\textbf{(3) Projection and outputs.} RANSAC fits an affine transform to MINIMA-LoFTR's matches. The mask extracted with SAM3 is projected into WGS84 through that affine transformation, yielding a candidate GeoJSON.

\paragraph{Tool 3: Commit results.}
The Worker's prompt (\Cref{appendix:prompts}) instructs the agent to commit results based on a specific policy over these three signals:  the RANSAC inlier count; a \emph{scale-consistency} score in $[0,1]$ measuring how closely the recovered affine's average scale matches the printed scale; and a \emph{road-name agreement} score in $[0,1]$ measuring the fraction of road names that actually appear at the matched location. The exact policy can be found in the worker prompt in \Cref{appendix:prompts}.

\paragraph{Tool 4: District shortcut.}
For documents the Reader flagged as covering an entire district, the Worker bypasses visual positioning and calls \texttt{lookup district}, which returns the named district's polygon directly from OS BoundaryLine~\cite{os_boundary_line}. This is faster and more precise than going through the whole pipeline.

\subsection{Critic}
\label{subsection:critic}

To test whether a second pass of visual verification can increase performance, we add an optional Critic stage. After the Worker submits its result, a separate LLM call with fresh context receives visual panels (\Cref{fig:critic-panel}) for the top-3 candidates by RANSAC inlier count including the final candidate the Worker committed, along with each candidate's match-quality signals (inlier count, scale consistency, road-name agreement). The Critic returns an action $a \in$ \{\texttt{approve}, \texttt{switch}, \texttt{retry locate}\}. \texttt{approve} accepts the candidate the Worker committed and ends the pipeline; \texttt{switch} switches the final candidate to another one of the top-3 that the Worker did not commit; \texttt{retry locate} re-invokes the Worker with a templated feedback string, instructing the Worker to find a better map centre. The Critic is allowed at most two such re-localisations per case.

\section{Experiment}
\label{section:results}

We evaluate GeoPlanAgent on all 208 cases of Plan2Map. The Reader, Worker, and Locate sub-agent all run on Gemini~3~Flash at temperature~0 for better reproducibility. All inference runs on a single Apple M3 Max laptop (40-core GPU, 128\,GB unified memory) with \texttt{fp16} autocast on MPS. SAM~3 fine-tuning was performed on an NVIDIA H200 SXM GPU (see \Cref{subsection:sam3-finetune}).

\paragraph{Evaluation Metrics}
Let $G$ denote the reference GeoJSON and $P$ the predicted GeoJSON. We evaluate each prediction with the following metrics. (1) \textit{IoU.} We compute geometric intersection-over-union,
$\mathrm{IoU}=|G\cap P|/|G\cup P|$, to jointly measure boundary tracing and geographic positioning. We report mean IoU, median IoU, $\%_{\mathrm{IoU}>0}$, and $\%_{\mathrm{IoU}\geq 0.8}$. The last metric captures high-fidelity recoveries and is inspired by the EXTRACT system of \citet{colebourn2025extract}. (2) \textit{Err (m).} This is the median centroid distance, in metres, between $G$ and $P$, designed to isolate localisation error from boundary-shape quality. We report the median because a small tail of severe localisation failures right-skews the distribution. (3) \textit{Acc@0.1D.} Following \citet{colebourn2025extract}, we report the fraction of cases whose centroid error is at most 10\% of the reference boundary diameter. We operationalise diameter as the Feret diameter, i.e., the longest pairwise distance between convex-hull vertices. This normalises localisation error by boundary size. (4) \textit{Cost and time.} Cost is the mean USD API cost per document at OpenRouter prices on 2026-05-24, summed over all LLM calls in the pipeline. Time is the mean wall-clock runtime per case on the laptop described above.




\begin{table*}[t]
\centering
\vspace{-10mm}
\footnotesize
\setlength{\tabcolsep}{2pt}
\renewcommand{\arraystretch}{1.10}
\begin{tabular}{@{}l c cccc cc cc@{}}
\toprule
\rowcolor{headerbg} & \textbf{Scope} & \multicolumn{4}{c}{\textbf{GeoJSON quality}} & \multicolumn{2}{c}{\textbf{Localization}} & \multicolumn{2}{c}{\textbf{Cost}} \\
\cmidrule(lr){3-6} \cmidrule(lr){7-8} \cmidrule(lr){9-10}
\textbf{Method} & \textbf{N} & \textbf{$\%_{\mathrm{IoU}>0}\uparrow$} & \textbf{Mean IoU$\uparrow$} & \textbf{Med IoU$\uparrow$} & \textbf{$\%_{\mathrm{IoU}\geq 0.8}\uparrow$} & \textbf{Err\,(m)$\downarrow$} & \textbf{Acc@0.1D$\uparrow$} & \textbf{\$/doc$\downarrow$} & \textbf{Time\,(s)$\downarrow$} \\
\midrule
\multicolumn{10}{@{}l}{\textcolor{sectiongray}{\emph{VLM end-to-end --- 40-case stratified subset (\Cref{appendix:vlm-subset})}}} \\
\rowcolor{white} Gemini-3-Flash   & 40  & 30.0\% & 0.053 & 0.000 & 0.0\% & 920  & 2.5\%  & \textbf{0.003} & \textbf{5}  \\
\rowcolor{white} Gemini-3.1-Pro   & 40  & 42.5\% & 0.112 & 0.000 & 0.0\% & 490  & 7.5\%  & 0.106 & 75 \\
\rowcolor{white} Claude-Opus-4.7  & 40  & 22.5\% & 0.044 & 0.000 & 0.0\% & 1131 & 0.0\%  & 0.059 & 7  \\
\rowcolor{white} GPT-5.5-Pro      & 40  & 50.0\% & 0.106 & 0.005 & 0.0\% & 386  & 10.0\% & 2.855 & 650 \\
\addlinespace[2pt]
\multicolumn{10}{@{}l}{\textcolor{oursgreen}{\textsc{\emph{Ours --- 40-case stratified subset}}}} \\
\rowcolor{oursbg} GeoPlanAgent    & 40  & 85.0\% & 0.721 & 0.901 & 67.5\% & 6.7 & 80.0\% & 0.043 & 162 \\
\addlinespace[2pt]
\midrule
\multicolumn{10}{@{}l}{\textcolor{sectiongray}{\emph{VLM end-to-end --- full dataset}}} \\
\rowcolor{white} Gemini-3.1-Pro   & 208 & 40.4\% & 0.108 & 0.000 & 1.4\% & 480 & 9.6\% & 0.106 & 75 \\
\addlinespace[2pt]
\midrule
\multicolumn{10}{@{}l}{\textcolor{oursgreen}{\textsc{\emph{Ours --- full dataset}}}} \\
\rowcolor{oursbg}       GeoPlanAgent  & 208 & 89.4\% & 0.736 & 0.904 & \textbf{67.8\%} & \textbf{4.6} & \textbf{78.8\%} & \textbf{0.043} & 153 \\
\rowcolor{oursbgstrong} \quad + Critic & 208 & \textbf{89.9\%} & \textbf{0.740} & \textbf{0.906} & \textbf{67.8\%} & \textbf{4.6} & \textbf{78.8\%} & 0.045 & 155 \\
\addlinespace[2pt]
\midrule
\multicolumn{10}{@{}l}{\textcolor{sectiongray}{\textsc{\emph{Other variants --- full dataset}}}} \\
\rowcolor{variantbg} Collapsed Reader & 208 & 88.9\% & 0.733 & 0.904 & 67.3\% & 4.7 & 78.4\% & 0.053 & 156 \\
\bottomrule
\end{tabular}
\caption{Main results on Plan2Map. \emph{VLM end-to-end} rows report direct GeoJSON prediction from frontier vision-language models with no tool access and reported on a 40-case stratified subset; we also report the GeoPlanAgent results on this subset. Gemini-3.1-Pro, the strongest VLM by mean IoU, is additionally run on the full 208 cases.}
\label{tab:main-result}
\vspace{-5mm}
\end{table*}

\subsection{Main Results}

\Cref{tab:main-result} reports the main results. On the full Plan2Map benchmark, GeoPlanAgent reaches 0.736 mean IoU and 0.904 median IoU, with 67.8\% of cases at IoU $\geq 0.8$. The median centroid error is 4.6\,m and \emph{Acc@0.1D} reaches 78.8\%, indicating that successful cases are both well traced and well localized. Direct VLM-to-GeoJSON baselines are substantially weaker: the strongest full-dataset direct baseline, Gemini-3.1-Pro, achieves only 0.108 mean IoU and 9.6\% Acc@0.1D.

For external context, EXTRACT is the closest deployed point of reference, but it is not a directly comparable baseline: it is evaluated on a private corpus and does not report the same full set of metrics \citep{colebourn2025extract}. We therefore use it only to calibrate the operating regime. EXTRACT reports 82\% Acc@0.1D at approximately \$0.12 per document, while GeoPlanAgent reaches 78.8\% Acc@0.1D on Plan2Map at \$0.043 per document under our stated Feret-diameter definition. The optional Critic gives a small gain (+0.003 mean IoU) at negligible extra cost, so we treat it as a lightweight verification variant rather than a load-bearing component.


\subsubsection{Direct VLM Baseline}
\label{subsection:abl-vlm-direct}

We compare GeoPlanAgent against frontier vision-language models that predict GeoJSON directly from the planning PDF, with no access to tools. The prompt (\Cref{fig:vlm-direct-prompt}) asks the model to read the document, identify the affected area, localize it, trace the boundary, and return real-world coordinates, compressing the full Plan2Map pipeline into a single multimodal call. \Cref{tab:main-result} reports 40-case stratified-subset results for Gemini-3-Flash~\citep{google2025geminiflash3}, Gemini-3.1-Pro~\citep{google2026gemini31pro}, Claude Opus-4.7~\citep{anthropic2026claudeopus47}, and GPT-5.5-Pro~\citep{openai2026gpt55}. We additionally run Gemini-3.1-Pro on all 208 cases because it gives the strongest cost-effective direct baseline on the subset.


\paragraph{Temperature.}
For the agentic pipeline, we set the temperature to 0 to improve the reproducibility of tool selection, schema-constrained outputs, and commit decisions. For the VLM-direct baselines, we instead use each provider's recommended default temperature of 1: at temperature 0, approximately 20\% of Gemini-3.1-Pro cases produced empty or malformed pixel-coordinate polygons or exhausted the output budget during reasoning. This choice gives the direct baseline a valid operating configuration rather than penalising it for a decoding setting that degrades Gemini thinking models. To keep the VLM-direct baseline comparable, we also chose a temperature of 1 for the other models.

\begin{figure}[h]
\centering
\includegraphics[width=\linewidth]{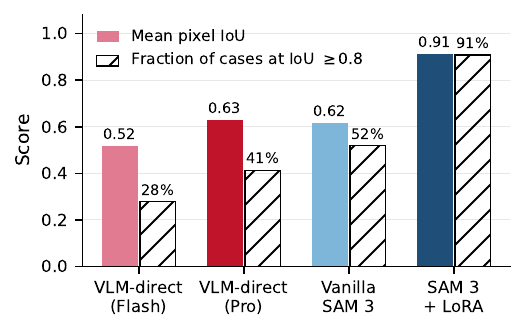}
\caption{Boundary segmentation, scored against hand-annotated map masks on all 208 cases. Solid bars: mean pixel IoU. Hatched bars: fraction of cases at IoU $\geq 0.8$. SAM 3 + LoRA leads by $\geq 0.30$ IoU and $\geq 39$ percentage points over the vanilla SAM 3 baseline.}
\label{fig:abl-finetune}
\vspace{-5mm}
\end{figure}

\paragraph{Results.}
All four VLM-direct models sit in the same band: mean IoU 0.04--0.11, none reaches IoU\,$\geq$\,0.8 on any case in the 40-subset, and median centroid errors range from 386\,m to 1.1\,km. On the same 40 cases our pipeline reaches mean IoU 0.721, a $6\times$ improvement on the matched subset. Of the four, Gemini-3.1-Pro and GPT-5.5-Pro are the top two and trade off complementarily: Gemini-3.1-Pro leads on mean IoU (0.112 vs 0.106); GPT-5.5-Pro has lower centroid error on centroid distance (median 386\,m vs 490\,m) and on \emph{Acc@0.1D} (10.0\% vs 7.5\%). We scale Gemini-3.1-Pro to the full 208-case set rather than GPT-5.5-Pro for two reasons: (i) Gemini leads on the mean-IoU metric reported in the main table; (ii) GPT-5.5-Pro's cost is $\approx 27\times$ higher per document (\$2.86 vs \$0.11), so a full 208-case run would cost $\approx \$600$ for GPT-5.5-Pro versus $\approx \$22$ for Gemini-3.1-Pro. The bulk of GPT-5.5-Pro's cost is internal reasoning tokens that do not translate into the geospatial-grounding metric, illustrating that reasoning depth alone does not close the gap to our tool-augmented multi-agent pipeline.

\subsubsection{Collapsed Reader}
\label{subsection:abl-collapsed-reader}
The \emph{Collapsed Reader} row of \Cref{tab:main-result} removes the dedicated Reader phase (\Cref{subsection:abl-collapsed-reader}): The Worker keeps its original role described above and is additionally responsible for extracting structured information from the PDF as well. Mean IoU is within noise of the separated baseline (0.733 vs.\ 0.736; median 0.904 / 0.904), but per-document cost rises 23\% from \$0.043 to \$0.053 because the worker now carries the PDF binary in conversation history across every tool call ($+63\%$ tokens per case). The dedicated Reader phase therefore buys token efficiency rather than accuracy, and furthermore allows us to configure different models for different stages of the pipeline.


\subsection{Ablations}
\label{section:ablations}

The ablation study focuses on two main components of the pipeline. \Cref{subsection:abl-finetune} (\emph{segmentation}) and \Cref{subsection:abl-locate} (\emph{localization}) score the two components in isolation, against per-pixel and per-coordinate ground truth, so each module is judged on its own terms rather than only on the final GeoJSON IoU.

\subsubsection{Boundary segmentation}
\label{subsection:abl-finetune}

This section isolates boundary segmentation from positioning by scoring against human-annotated reference masks in pixel space. \Cref{fig:abl-finetune} compares four boundary segmentation regimes on pixel-level IoU. \emph{SAM3-LoRA (ours)} is scored under the same 5-fold leave-one-out setup used by the main pipeline at inference time (\Cref{subsection:worker}); per-fold numbers are in \Cref{tab:sam3-cv}. Since we use SAM3's feature to segment boundaries via text prompts, for the results of \emph{Vanilla SAM~3} (base SAM~3, no fine-tune), we first sweep five candidate prompts and report the best (full sweep in \Cref{subsection:abl-vanilla-sam-prompt-search}). For \emph{SAM3-LoRA (ours)}, we used the prompt it was fine-tuned on, which is ''planning boundary''. \emph{VLM-direct} is Gemini-3-Flash and Gemini-3.1-Pro tracing the polygon directly from the map image. We select these two models since Flash is the same model our pipeline runs on (which allows us to isolate the contribution of tool use versus a single-call solution holding the backbone fixed) and since Gemini-3.1-Pro is the next-stronger member of the same family, showing what the next-stronger member within the same model family can recover without tool-augmentation. Overall, we find that the gap in mean IoU between vanilla SAM~3 and our fine-tuned adapter is 0.30, which quantifies the contribution of the supervised fine-tune. We do not treat this as a head-to-head comparison with EXTRACT\citep{colebourn2025extract}, which is evaluated on a private corpus; instead, the result shows that boundary segmentation is no longer the dominant bottleneck once adapted to Plan2Map.


\begin{figure*}[t]
\centering
\vspace{-15mm}
\resizebox{0.95\textwidth}{!}{
\includegraphics[]{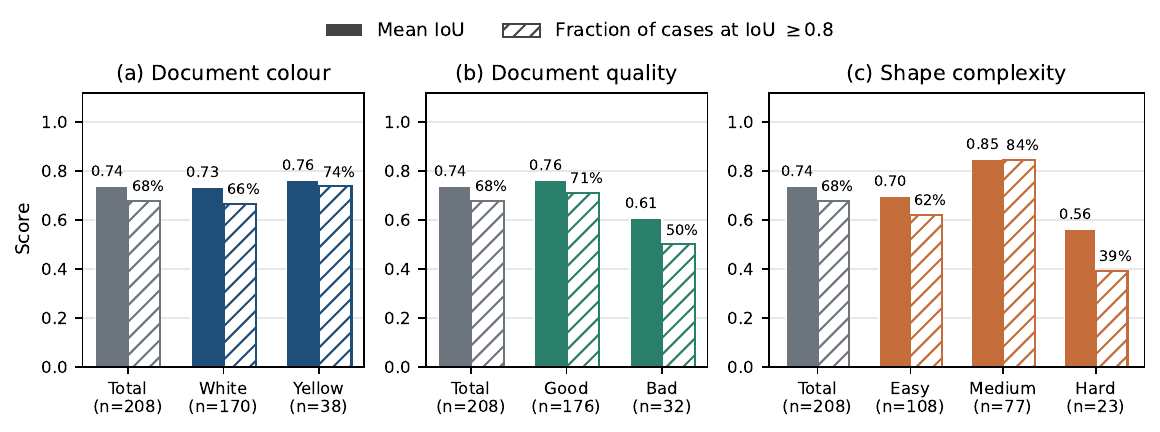}}
\caption{Mean IoU (filled) and fraction of cases at IoU\,$\geq$\,0.8 (hatched), from left to right, broken down by colour, quality, and shape complexity. The \textit{Total} bar is taken from \Cref{tab:main-result}.}
\label{fig:cls-bars}
\vspace{-5mm}
\end{figure*}

\begin{table}[h]
\centering
\setlength{\tabcolsep}{3pt}
\renewcommand{\arraystretch}{0.95}
\footnotesize
\begin{tabular}{@{}lcrrr@{}}
\toprule
\textbf{Configuration} & \textbf{N} & \textbf{Err\,(m)$\downarrow$} & \textbf{$\%_{<500\mathrm{m}}\uparrow$} & \textbf{$\%_{<1\mathrm{km}}\uparrow$} \\
\midrule
Locate                  & 208 & 176        & 78.8\%          & 91.3\%          \\
Locate + 5 tools        & 208 & 181        & 82.2\%          & \textbf{92.8\%} \\
VLM-direct (Flash)      & 208 & 522        & 47.1\%          & 73.6\%          \\
VLM-direct (Pro)        & 208 & 455        & 53.4\%          & 78.8\%                 \\
\midrule
\rowcolor{oursbg} GeoPlanAgent  & 208 & \textbf{5} & \textbf{88.9\%} & 92.3\%          \\
\bottomrule
\end{tabular}
\caption{Locate-stage centroid error on all 208 cases. We compare the production Locate sub-agent using its single OS Open Names query tool, a variant with five additional geocoding tools, and two single-shot VLM-direct geocoders. The final row reports GeoPlanAgent after map-registration refinement.}
\label{tab:abl-locate}
\vspace{8mm}
\end{table}

\subsubsection{Localization}
\label{subsection:abl-locate}

We isolate localization by scoring predicted centroids against the reference polygon centroid. \Cref{tab:abl-locate} compares the production Locate sub-agent using OS Open Names, a variant with five additional geocoding tools, single-shot VLM geocoders, and the final GeoPlanAgent output after map-registration refinement. The Locate sub-agent reaches 176\,m median error and places 78.8\% of cases within 500\,m, substantially outperforming VLM-direct geocoding (522\,m and 47.1\% for Gemini-3-Flash). This shows that tool-backed geocoding provides a useful coarse location before visual registration.



Adding the five extra geocoders only nudges within-500\,m from 79\% to 82\% and slightly worsens the median (176 vs.\ 181\,m); the 6-tool variant does post the table's nominal best within-1\,km (92.8\% vs.\ 92.3\% for GeoPlanAgent), but this difference is not statistically relevant.

The bottom row of \Cref{tab:abl-locate} shows how well the sliding-window approach refines localization. The median centroid error drops from 176\,m to 5\,m - a $38\times$ tightening - and the within-500\,m fraction rises from 78.8\% to 88.9\%. Within-1\,km is essentially unchanged (91.3\% to 92.3\%) because the Locate centre was already inside the kilometre band on most cases; the gain from sliding-window registration is concentrated in the sub-500\,m precision regime that the Locate stage alone cannot reach.


\subsection{Further Analysis}

\Cref{fig:cls-bars} breaks down GeoPlanAgent performance by document colour, document quality, and shape complexity, as defined in \Cref{sec:dataset}. Performance is relatively stable across document colour, but degrades on poor-quality scans and hard geometries. This suggests that the remaining failures are less driven by superficial visual style and more by robust map interpretation under low document quality and complex or multi-part boundaries.

\section{Discussion and Conclusion}

We introduced \textbf{Plan2Map}, a public benchmark of 208 manually reviewed UK Article~4 Direction records that pair source planning PDFs with verified GeoJSON boundaries and rendered location-map visualisations. We framed planning-record digitisation as an indirect geospatial reconstruction task, where the target boundary must be recovered from textual, visual, and geographic evidence rather than read from an explicit field in the PDF. To address this benchmark, we proposed \textbf{GeoPlanAgent}, a decomposition-based system that mirrors the task structure through a Reader, a Locate sub-agent backed by an offline UK geocoder, a Worker that registers the planning map and segments the boundary, a projection step, and an optional Critic.

On the 208-case benchmark, GeoPlanAgent with Critic reaches 0.740 mean IoU, 0.906 median IoU, and 67.8\% of cases at IoU $\geq 0.8$. Component ablations show that direct VLM-to-GeoJSON prediction remains unreliable, that LoRA fine-tuning is important for map-boundary segmentation, and that localisation and map registration remain the main bottlenecks. Together, Plan2Map and GeoPlanAgent provide a testbed for studying document-grounded geospatial reconstruction, with future work focused on more robust localisation, stronger verification in borderline cases, and extension to additional planning domains and jurisdictions.

\clearpage

\section*{Limitations}
Plan2Map currently focuses on the UK, where the richness of open geospatial infrastructure (Code-Point Open, OS Open Names, OS BoundaryLine, OS OpenMap Local, OS Open Zoomstack) enabled us to build and rigorously evaluate a complete end-to-end pipeline. The agent architecture and task framing are designed to be jurisdiction-agnostic; extending to other countries requires only re-pointing each tool at the local geospatial equivalent, making Plan2Map a concrete template for analogous efforts in other countries.

The boundary-segmentation step assumes that the application boundary is visually drawn on the planning map. Documents whose application area is described purely by text (e.g. ``land north of 14 Manor Road'') with no overlaid boundary line fall outside the pipeline's competence\textemdash the Locate sub-agent will still produce a centre, but SAM~3 cannot segment a polygon that does not exist on the page.

\section*{Data Attribution}
Contains OS data \textcopyright{} Crown copyright and database right 2026.
Contains Royal Mail data \textcopyright{} Royal Mail copyright and database right 2026.
Contains National Statistics data \textcopyright{} Crown copyright and database right 2026.
Source planning documents and reference GeoJSON boundaries are reproduced from planning.data.gov.uk under the Open Government Licence v3.0.

\section*{Acknowledgement}
We would like to thank Jamie Heagerty from the Number 10 and Liam Wilkinson from the Cabinet Office for the useful discussions.

\bibliography{custom}

\clearpage

\appendix
\title{Appendix}

\section{Dataset Details}
\label{appendix:dataset-details}

We curated a benchmark of 208 UK Article~4 Direction records. Article~4 Directions are legal instruments through which UK local planning authorities withdraw nationally granted permitted development rights for specified areas, typically to protect conservation areas, restrict change-of-use, or preserve landscape character \citep{mhclg2024planningpracticeguidance}. Each released case bundles the original regulatory PDF, a machine-readable reference boundary in GeoJSON, and a rendered location-map visualisation of the boundary on an OpenStreetMap basemap. To our knowledge, this is the first publicly released corpus that links UK planning-control documents to verified, machine-readable boundary geometries, making it suitable for evaluating systems that must read regulatory text, interpret an associated map, and recover a GIS-ready representation of the planning boundary. The records span 1958--2025 and cover twenty-nine distinct local planning authorities across England. The corpus is drawn exclusively from UK planning material and depends on UK gazetteers and base mapping, so it should be treated as a UK-specific benchmark. The metadata and the 208 data samples are anonymously available \href{https://osf.io/8d4nm/overview?view_only=35ec93320491426f86b2f4e07ed42153} {here}.

\subsection{Sources and Construction}

Raw source data were obtained from open UK planning datasets published under the Open Government Licence. Each raw entry consists of an Article~4 Direction PDF paired with a corresponding GeoJSON polygon retrieved from the same authority record. We then carried out a six-stage curation workflow with quality-control checkpoints at each transition: (i)~\emph{document acquisition} from authorised sources; (ii)~\emph{quality assessment} covering visual inspection, completeness, and authenticity; (iii)~\emph{boundary visualisation}, in which the supplied GeoJSON polygon is rendered on an OpenStreetMap basemap to produce the per-case location-map PNG and to enable side-by-side comparison against the boundary drawn in the PDF; (iv)~\emph{metadata annotation} of thirteen structured fields; (v)~\emph{validation} by visual shape matching and cross-reference against the source record; and (vi)~\emph{final quality control} through peer review and discrepancy documentation. The pipeline reduced 270 raw entries to 208 release-ready cases. Specifically, we removed 40 records whose PDF boundary and supplied GeoJSON disagreed (e.g., due to shape mismatches, split maps, or polygons covering only part of a multi-map PDF), 9 records whose PDFs contained no boundary map, and 7 duplicates; we further merged 5 multi-GeoJSON groups (consolidating 11 source rows into 5 cases) where the same PDF was distributed across several authority entries.

\subsection{Dataset Composition}
 Each released data item includes three artefacts. The \textbf{PDF} contains the regulatory text, schedule of withdrawn permitted development rights, boundary map, and authority signatures. The \textbf{GeoJSON} is a single feature carrying a \texttt{Polygon} or \texttt{MultiPolygon} geometry in WGS84, representing the application boundary. The \textbf{location-map PNG} overlays the GeoJSON on OpenStreetMap with WGS84 axes for human-readable inspection.
 
The accompanying metadata table records, for each case, an administrative area (county or local authority), a free-text site description, a categorical boundary shape, a document-quality label, a shape-complexity label, and a validation flag. Boundary-shape categories are \emph{rectangular}, \emph{quadrilateral}, \emph{triangular}, \emph{irregular polygon}, \emph{L-shaped}, \emph{elongated strip}, \emph{multiple separate polygons}, and \emph{multiple complex irregular polygons}. Document quality is annotated as \emph{good} for clear text and legible maps, and \emph{bad} for poor scans, faded text, or unclear boundaries. Shape complexity is annotated as \emph{easy} (fewer than eight vertices), \emph{medium} (8--20), or \emph{hard} (more than twenty, or multiple disjoint components). The ``Shape Matches'' flag, with values \{\emph{Yes}, \emph{Yes--almost}\}, records whether the supplied GeoJSON visually matches the boundary drawn in the PDF; 96.6\% of released cases are exact matches and 3.4\% near matches, with all non-matching cases removed at curation. Identification fields maintain referential integrity between the PDF, the GeoJSON, and the metadata table (\Cref{Unique_Identification} ).

\subsection{Distribution}
 
The dataset spans 1958--2025 with a median document year of 1997; cases are distributed across the 1970s (20.2\%), 1980s (18.3\%), 1990s (12.5\%), 2000s (17.3\%), 2010s (25.0\%), and 2020s (4.8\%), with the remaining 1.9\% pre-1970. Geographically, the corpus covers eighteen counties or county-level administrative areas, concentrated in Kent (16.8\%), Norfolk (15.4\%), Hertfordshire (14.9\%), and London (13.9\%), with the remaining 38.9\% distributed across fourteen further areas, including Cambridgeshire, Staffordshire, Bristol, Leicestershire, Surrey, and Lancashire. At the local-authority level, the five largest groupings are St~Albans (28), Dover District (25), South Norfolk (23), South Staffordshire (13), and Bristol (12). Authority frequency reflects publication practices in the underlying datasets rather than a systematic geographic stratification.
 
\Cref{tab:data-characteristics} summarises the document-level distributions. The combination of multi-decade scans, mixed quality, and varied shape complexity provides a non-trivial benchmark for end-to-end planning-record digitisation and a usable signal for stratified evaluation.
 
\begin{table}[t]
\centering
\setlength{\tabcolsep}{1.5pt}
\small
\caption{Distribution of dataset characteristics across the 208 released cases.}
\begin{tabular}{@{}lll@{}}
\toprule
\textbf{Field} & \textbf{Values} & \textbf{Distribution} \\
\midrule
Doc. Date     & 1958--2025         & Median 1997 \\
Doc. Colour   & White/Yellow & 82\% / 18\% \\
Doc. Quality  & Good / Bad         & 85\% / 15\% \\
Shape Complexity & Easy/Medium/Hard & 52\% / 37\% / 11\% \\
\bottomrule
\end{tabular}
\label{tab:data-characteristics}
\end{table}

\subsection{Stratified 40-case subset for VLM-direct baselines}
\label{appendix:vlm-subset}

To compare direct VLM-to-GeoJSON baselines without paying the cost of running every frontier model on all 208 cases, we construct a stratified 40-case subset over the \emph{Document Quality} $\times$ \emph{Shape Complexity} grid (six strata: good/bad $\times$ easy/medium/hard); the resulting allocation is shown in \Cref{tab:vlm-subset-strata}. Allocation per stratum guarantees a floor of two cases and otherwise distributes proportionally to stratum population. Sampling is deterministic (\texttt{seed=42}).

Quality and shape complexity are the metadata axes most likely to differentially stress the VLM-direct baseline --- quality drives map and label legibility, shape complexity drives boundary-tracing difficulty --- making them the natural axes to balance across. Stratifying additionally on colour would create 18 cells from only 40 cases, several of which would contain 0--1 cases in the joint distribution.

\begin{table}[h]
\centering
\footnotesize
\setlength{\tabcolsep}{6pt}
\renewcommand{\arraystretch}{1.05}
\begin{tabular}{@{}lcc@{}}
\toprule
\textbf{Stratum} & \textbf{Allocated} \\
\midrule
good $\times$ easy   & 15  \\
good $\times$ medium & 12 \\
good $\times$ hard   &  3 \\
bad  $\times$ easy   &  4 \\
bad  $\times$ medium &  3 \\
bad  $\times$ hard   &  3 \\
\midrule
Total & 40 \\
\bottomrule
\end{tabular}
\caption{Stratum composition of the 40-case VLM-direct comparison subset.}
\label{tab:vlm-subset-strata}
\end{table}

\section{Data Collection Methodology}


The documents and associated boundary files were collected from UK planning authority sources\footnote{https://www.planning.data.gov.uk/entity/?dataset=article-4-direction-area}. Processing followed a six-stage workflow: document acquisition, quality assessment, boundary extraction from the available GeoJSON data, metadata annotation, validation, and final quality control. At each stage, records were checked for completeness, PDF-GeoJSON consistency, and discrepancies between the boundary drawn in the source document and the corresponding vector geometry.

Source URLs, metadata, and release scripts will be made available with the final public release.

\begin{table}[h]
\centering
\caption{Identification Metadata Fields}
\begin{tabular}{@{}ll@{}}
\toprule
\textbf{Field} & \textbf{Description} \\ \midrule
Sl no & Sequential (1--208) \\
Folder ID & Path to Folder \\
Geojson ID & Path to GeoJSON \\ \bottomrule
\end{tabular}
\label{Unique_Identification}
\end{table}

\begin{table*}[h]
\centering
\caption{Geographic Metadata Fields}
\begin{tabular}{@{}lll@{}}
\toprule
\textbf{Field} & \textbf{Description} & \textbf{Example} \\ \midrule
County & Administrative area & Kent, London (Westminster) \\
Exact Location & Site description & ``Land at Golgotha, Shepherdswell'' \\
Boundary Shape & Geometry type & Irregular polygon, Rectangular \\ \bottomrule
\end{tabular}
\end{table*}

\section{Inference Pipeline}
\label{sec:impl-appendix}

This appendix expands the per-component description in \Cref{section:geoplanagent}, in the order each component executes at inference time: PDF reading, rotation correction, geographic localisation, basemap tile preparation, sliding-window map matching, and the district shortcut for entire-area documents.

\subsection{PDF Reading and Page Rendering}
\label{appendix:text-extraction}

The Reader (\Cref{subsection:reader}) consumes the PDF as a single multimodal LLM call: the raw PDF bytes are attached inline to the message and Gemini~3~Flash extracts the structured record of spatial evidence in one shot. \Cref{tab:pdfinfo-schema} lists the full set of fields; together they cover the spatial identifiers a downstream geocoder might use, the named features visible on the map image, and per-page routing metadata that lets the Worker pick the right map page without re-reading the document. We use no separate text-extraction or OCR cascade --- the model reads the embedded text layer, the rasterised pages, and any embedded form fields directly. The Reader runs at temperature~0 with a two-retry schema-validation budget, and a strict output validator rejects clearly degenerate outputs (e.g.\ all named-feature fields empty while spatial-identifier fields are populated, indicating a partial generation failure).

For positioning, every page that the Reader marks as a positionable map is rasterised at 200\,DPI with PyMuPDF and passed to the auto-rotation classifier (\Cref{subsection:rotation-classifier}) before being handed to the Worker.

\begin{table*}[h]
\centering
\setlength{\tabcolsep}{4pt}
\renewcommand{\arraystretch}{1.05}
\small
\caption{Fields the Reader populates per planning document. Fields are grouped by role; downstream stages consume different subsets. A few fields hold the same underlying signal in different forms (raw vs.\ parsed, image- vs.\ text-derived) to suit different downstream consumers.}
\label{tab:pdfinfo-schema}
\begin{tabular}{@{}lp{0.78\textwidth}@{}}
\toprule
\textbf{Field} & \textbf{Description} \\
\midrule
\multicolumn{2}{@{}l}{\emph{Spatial identifiers from the document text.}} \\
\texttt{site\_address}              & Full site address (location of the boundary). Council, agent, and architect addresses are ignored. \\
\texttt{postcodes}                  & UK postcodes found anywhere in the text. \\
\texttt{grid\_refs}                 & OS British National Grid references. \\
\texttt{house\_number\_road\_pairs} & House-numbered addresses in the form \texttt{<numbers> <road>}; ranges collapsed (e.g.\ \texttt{126--134 Norwich Road}). \\
\texttt{parish\_names}              & Civil or ecclesiastical parishes named in the text. \\
\texttt{admin\_region}              & Most specific administrative region (district or borough); district/borough preferred over county. \\
\midrule
\multicolumn{2}{@{}l}{\emph{Named features visible on the map image.}} \\
\texttt{road\_names}                & Roads named in the document text or on the map image. \\
\texttt{place\_names}               & Named places --- villages, neighbourhoods, landmarks, named buildings. \\
\texttt{visible\_map\_labels}       & Labels printed on the map image itself, copied verbatim (separate from text-body content). \\
\texttt{adjacency\_hints}           & Features bordering the boundary, extracted from phrases like ``adjoining $X$'', ``bounded by $Y$''. \\
\midrule
\multicolumn{2}{@{}l}{\emph{Geographic context.}} \\
\texttt{likely\_town\_or\_city}     & Best single town or city synthesised from all signals; disambiguates homonymous road names. \\
\texttt{scale}                      & Printed map scale denominator (e.g.\ ``1:2500''). \\
\midrule
\multicolumn{2}{@{}l}{\emph{District shortcut control.}} \\
\texttt{is\_district\_wide}         & True if the boundary covers an entire borough, district, ward, parish, or conservation area. \\
\texttt{district\_name}             & Canonical UK admin name with optional ``$\mid$''-separated alternates (e.g.\ ``City of Westminster, UK $\mid$ Westminster, UK''). \\
\midrule
\multicolumn{2}{@{}l}{\emph{Page-level routing.}} \\
\texttt{map\_pages}                 & Ranked list of positionable map-page indices. \\
\texttt{map\_page\_details}         & Per-page metadata for every page that contains map-like content: page index, category (positionable or discarded), area group (equivalence class over views of the same site), boundary clarity, detail level (close / medium / wide), and a short caption. \\
\bottomrule
\end{tabular}
\end{table*}

\subsection{Auto-Rotation Classifier}
\label{subsection:rotation-classifier}

Many UK planning maps are laid out sideways on the page. We therefore employ a learned 4-class ResNet50 classifier over $\{0^\circ, 90^\circ, 180^\circ, 270^\circ\}$, where the class is the clockwise rotation needed to make the page north-aligned. Training data is hand-annotated over the 208 cases in the dataset: each map is labelled with a corrective rotation $R \in \{0, 90, 180, 270\}$, and each case generates four training samples by applying every $k \in \{0, 90, 180, 270\}$ to the map and labelling the result with $(R - k) \bmod 360$. This yields a class-balanced pool regardless of the heavy $0^\circ$ skew in the raw labels.

\paragraph{Cross-validation and leakage control.}
Because the training pool overlaps with the evaluation set, we train as 5-fold cross-validation using the same fold-by-case assignment as SAM~3 (\Cref{subsection:sam3-finetune}). At inference, each evaluation case is routed to the fold whose checkpoint did \emph{not} see it during training, so no case is ever rotated by a model that was trained on its label.

\paragraph{Training hyperparameters.}
\Cref{tab:rotation-hparams} lists the full configuration. Horizontal flip is omitted as augmentation because it would invert the rotation label.

\paragraph{Inference.}
To reduce single-view errors during inference, we apply 4-rotation test-time augmentation (TTA): the classifier is run on the input and its three $90^\circ$, $180^\circ$, and $270^\circ$ clockwise-rotated copies, giving four predictions of the same underlying orientation. Each prediction is realigned to the original frame --- a predicted rotation $c$ on a $k$-rotated input corresponds to $(c + k) \bmod 360$ in the original --- and the four resulting softmaxes are averaged. We abstain (return $0^\circ$) whenever the averaged top-class probability is below 0.50; this converts the residual error mode into ``no rotation'' rather than ``wrong rotation''. We do this because if the map is wrongly rotated, the rest of the pipeline will fail; abstaining and leaving the map upright is less destructive than rotating it incorrectly. Across folds the classifier reaches mean top-1 accuracy $0.981 \pm 0.010$ with TTA, compared to $0.953 \pm 0.060$ without; the gain is concentrated on the worst fold (\Cref{tab:rotation-cv}).

\begin{table}[h]
\centering
\footnotesize
\setlength{\tabcolsep}{4pt}
\renewcommand{\arraystretch}{1.1}
\caption{Auto-rotation classifier training configuration.}
\label{tab:rotation-hparams}
\begin{tabular}{@{}l p{0.58\columnwidth}@{}}
\toprule
\textbf{Setting} & \textbf{Value} \\
\midrule
Backbone            & ResNet50 (ImageNet V2), full fine-tune \\
Input resolution    & $768 \times 768$ \\
Random resized crop & scale $(0.6, 1.0)$, aspect $(0.85, 1.18)$ \\
Colour jitter       & brightness/contrast 0.3, saturation 0.15, hue 0.05 \\
Random erasing      & $p{=}0.4$, scale $(0.02, 0.18)$ \\
Optimiser           & AdamW, lr $10^{-4}$, weight decay $10^{-4}$ \\
LR schedule         & Cosine \\
Gradient clipping   & norm 1.0 \\
Batch size          & 16 \\
Epochs              & 20 max, patience 4 \\
\bottomrule
\end{tabular}
\end{table}

\begin{table}[h]
\centering
\small
\setlength{\tabcolsep}{6pt}
\begin{tabular}{@{}lccc@{}}
\toprule
\textbf{Fold} & \textbf{$|V|$} & \textbf{No TTA} & \textbf{TTA} \\ \midrule
0    & 43 & 0.837 & 1.000 \\
1    & 40 & 0.975 & 0.975 \\
2    & 42 & 1.000 & 0.976 \\
3    & 41 & 0.951 & 0.976 \\
4    & 42 & 1.000 & 0.976 \\ \midrule
Mean & 208 & 0.953 & \textbf{0.981} \\
Std  &     & 0.060 & \textbf{0.010} \\ \bottomrule
\end{tabular}
\caption{Auto-rotation classifier 5-fold out-of-fold top-1 accuracy, with and without 4-rotation test-time augmentation. $|V|$ is the held-out validation size per fold. TTA improves the mean and standard deviation, at the cost of small drops on two of the folds that were already saturated.}
\label{tab:rotation-cv}
\end{table}

\subsection{Additional Geocoder Tools for the Locate Sub-Agent}
\label{appendix:locate-tools}
\label{subsection:locate-sub}

In \Cref{subsection:abl-locate} we test if the performance of the Locate sub-agent improves if it is given additional 5 offline geocoder tools at its disposal, instead of just one. In the following we will describe all tools --- the original tool of the Locate sub-agent and the additional 5 tools for the ablation --- in detail:

\begin{itemize}[leftmargin=*, itemsep=2pt, topsep=2pt]
  \item \textbf{Place-name lookup} --- searches OS Open Names for villages, hamlets, suburbs, named roads, churches, schools, hospitals, recreation grounds, named buildings, and other named features. Queries can be scoped to a named local authority for disambiguation. \emph{Default for main pipeline.}
\item \textbf{Postcode lookup} --- resolves a UK unit postcode to its centroid and admin district via Code-Point Open~\citep{code_point_open}.
  \item \textbf{Grid-reference parser} --- parses an OS British National Grid reference into a coordinate. Accepts two-letter-prefixed forms (e.g.\ ``TL 150 067'', ``TR3559'') as well as labelled easting/northing strings (e.g.\ ``485700 E 148600 N'').
  \item \textbf{Road-name lookup} --- returns one centroid per matching road instance from OS OpenMap Local~\citep{os_openmap_local}. Queries can be scoped to a named local authority.
\item \textbf{Road intersection} --- finds the intersections of 2--3 named roads in OS OpenMap Local.
\item \textbf{Local-authority containment check} --- tests whether a coordinate falls inside a named local-authority polygon from OS BoundaryLine~\citep{os_boundary_line}, flagging candidates geocoded outside the expected administrative area.
\end{itemize}

\subsection{Map-Tile Matching (Tool 2)}
\label{subsection:matching}
\label{subsection:tile-rendering}
\label{subsection:mask-postproc}

A single \texttt{match} call refines a coarse map-centre into a projected boundary polygon by aligning the rendered planning map against OS Open Zoomstack basemap tiles, segmenting the boundary, and projecting the resulting mask. The steps below describe each phase in detail.

\paragraph{Tile rendering.}
Tiles are rendered offline from the OS Open Zoomstack~\citep{os_open_zoomstack} GeoPackage by a styler that approximates the look of UK planning-map basemaps: pink-tinted buildings, white-filled grey-cased roads, light green woodland and greenspace, light blue surface water, dark grey railways, and a light cream background. Geometry is read in the UK's national grid coordinate system, reprojected to the Web-Mercator projection used by standard web map tiles, and rasterised to pixels.

\paragraph{Scale matching.}
For each candidate centre the planning map is resized so that one map pixel covers the same ground distance as one Web-Mercator tile pixel --- this is needed for MINIMA-LoFTR's feature matching to work well. We compute this by deriving the metres-per-pixel on each side and resizing the map to close any residual gap.

The map's metres-per-pixel is
\begin{equation*}
m_{\text{px}} = (25.4 / \text{DPI}) \cdot 10^{-3} \cdot s,
\end{equation*}
where $s$ is the printed scale denominator (e.g.\ $s=2500$ for a 1:2500 map at 200\,DPI): the factor $25.4/\text{DPI}$ converts pixels-per-inch into millimetres-per-pixel, the $10^{-3}$ converts millimetres to metres, and multiplying by $s$ scales page-metres up to ground-metres.

The tile's metres-per-pixel at latitude $\phi$ and zoom $z$ is
\begin{equation*}
t_{\text{px}}(\phi, z) = C \cos(\phi) / 2^z,
\end{equation*}
with $C \approx 156{,}543$\,m the ground distance per tile pixel at zoom~$0$ on the equator. The $\cos(\phi)$ factor accounts for Web-Mercator's horizontal stretch towards the poles, and the $2^z$ factor reflects the standard doubling of resolution per zoom level.

We then pick the closest integer zoom by solving $t_{\text{px}}(\phi, z) = m_{\text{px}}$ for $z$, giving
\begin{equation*}
z^{\star} = \mathrm{round}\bigl(\log_2(C \cos(\phi) / m_{\text{px}})\bigr),
\end{equation*}
clamped to $[15, 19]$ --- the band that covers typical UK planning-map scales (roughly 1:1250 to 1:25{,}000). Because the tile zoom is integer-valued, $t_{\text{px}}(\phi, z^{\star})$ generally will not exactly equal $m_{\text{px}}$; we close this residual gap by resizing the planning map by the factor $m_{\text{px}} / t_{\text{px}}(\phi, z^{\star})$ (area interpolation when downscaling, cubic when upscaling). If this factor falls outside $[0.3, 3.0]$, this scale configuration is dropped --- such extreme rescaling typically indicates that the Reader mis-extracted the printed scale.

\paragraph{Zoom and scale sweep.}
Because $z^{\star}$ is the rounded best zoom and the assumed metres-per-pixel can drift from reality (scanning artefacts in the source PDF distort the effective rendering DPI, and the extracted printed scale may itself be incorrect or not present), we do not rely on a single configuration: MINIMA-LoFTR is invoked at the best zoom and its two integer neighbours $\{z^{\star}-1, z^{\star}, z^{\star}+1\}$, plus two extra runs at $z^{\star}$ with $m_{\text{px}}$ perturbed by $\pm 15\%$. When the Reader did not extract a printed scale, the search instead sweeps six canonical UK planning-map scales (1:1250, 1:2500, 1:5000, 1:10\,000, 1:15\,000, 1:25\,000) at the zoom each implies, plus $\pm 15\%$ perturbations around the modal scale ($s=2500$). The tile canvas around the candidate is sized to cover the resized map plus a $\sigma$-driven margin, clamped to a $3{\times}3$ to $17{\times}17$ tile grid with odd dimensions on both axes so the canvas is symmetric around the candidate centre.

\paragraph{Sliding-window inner loop.}
For each of the above mentioned (centre, zoom, scale-perturbation) configurations the resized map is slid across the OS tile canvas with a step that targets $\approx 100$ windows per configuration: $\mathrm{step} = \max(32, \sqrt{(c_h - r_h)(c_w - r_w) / 100})$, where $c_h, c_w$ are the canvas's height and width in pixels and $r_h, r_w$ are the resized map's. The 32-pixel floor avoids redundant overlapping windows. At each window we run MINIMA-LoFTR and fit a 4-DOF similarity transform (rotation, uniform scale, translation) by RANSAC with a 10-pixel reprojection threshold and a fixed seed for reproducibility. Windows with fewer than 5 inliers are discarded; surviving windows are kept per configuration keyed by (centre, zoom), at most one candidate per configuration, with a global cap of 5. This diversity-bucketed top-$K$ scheme prevents one strong configuration from saturating every slot with near-duplicate windows.

\paragraph{Composite re-rank.}
The surviving top-$K$ candidates are re-ranked in two passes. First, each candidate's RANSAC inlier count $V$ is multiplied by a quadrant-coverage factor $Q/4$, where $Q \in \{0, 1, 2, 3, 4\}$ is the number of map quadrants containing at least one inlier; this penalises matches whose inliers cluster in one corner. Second, when the Reader extracted any road names, candidates are reweighted by $V \cdot (1 + r)^2$ where $r$ is the fraction of those names that appear within a 1500\,m radius of the candidate's predicted centre in the offline OS Open Zoomstack road graph. Candidates with no nearby OS roads (sparse rural cartography) receive a neutral $1.0$ multiplier. The top-scoring window after both passes is returned to the Worker as the candidate from this call.

\paragraph{Output signals.}
Beyond the raw inlier count, two consistency axes feed the Worker's commit policy. \emph{Scale-consistency} is $\min(s, 1/s)^2$ where $s$ is the average length of the first two columns of the recovered affine matrix ($s \approx 1$ means the pre-resize was correct). \emph{Road-name agreement} is the matched-to-total road-name ratio computed for the composite re-rank above, reported here to the Worker as an additional commit signal.

\subsection{Critic Panel Example}
\label{appendix:critic-panel}

\begin{figure}[h]
\centering
\centering
\includegraphics[width=\linewidth]{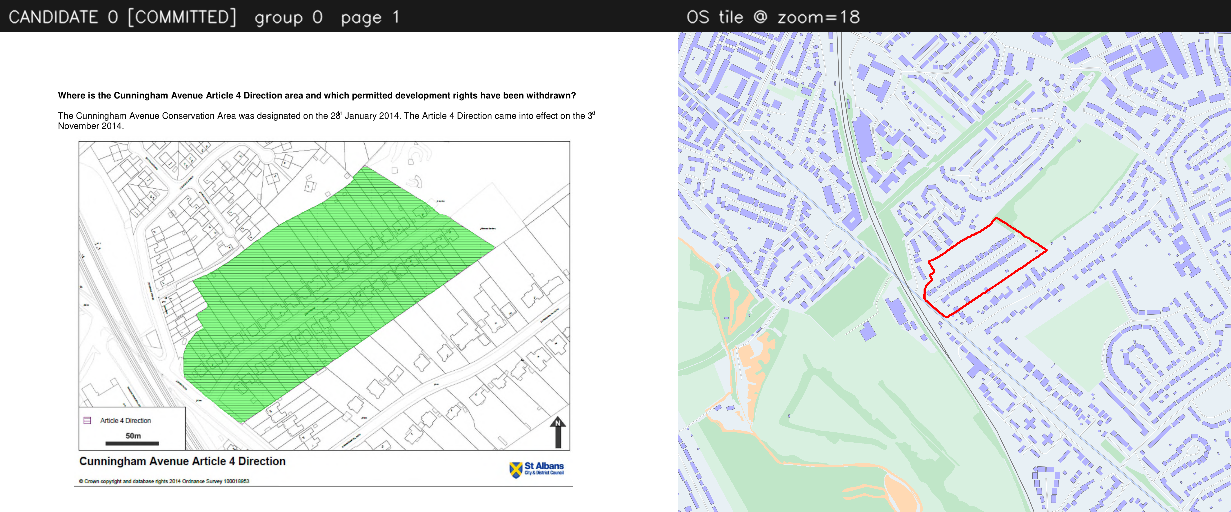}
\caption{Example of the visual input the Critic receives, here for a single stored candidate. \emph{Left:} the planning-map page with the SAM~3 mask overlaid in translucent green; a \texttt{[COMMITTED]} tag marks the Worker's chosen candidate. \emph{Right:} the OS Open Zoomstack tile render at the matched window, with the projected polygon outlined in red. When the Worker has considered multiple candidates, the top-3 by RANSAC inlier count are sent as separate panels; the Critic compares them and returns \texttt{approve}, \texttt{switch}, or \texttt{retry\_locate}.}
\label{fig:critic-panel}
\end{figure}

\section{SAM~3 Fine-Tuning}
\label{appendix:sam3}

\subsection{Architecture, Loss, and Optimisation}
\label{subsection:sam3-finetune}

SAM~3 is fine-tuned on hand-annotated boundary polygons for the 208 cases in the dataset; each annotation traces the boundary as drawn on the planning page. We fine-tune with rank-16 LoRA adapters on every $\{q, k, v, o\}$ attention projection and the two MLP projections of every transformer subsystem (vision encoder, text encoder, geometry encoder, DETR encoder/decoder, mask decoder). The weights of the mask embedder, presence head, and semantic projection layer are fully fine-tuned without LoRA. The training and inference query phrase is fixed to ``planning boundary''. \Cref{tab:sam3-hparams} lists the full training configuration.

SAM~3 has two output heads: a \emph{semantic} head that produces a per-pixel score map for the queried class, and an \emph{instance} head that produces object-instance proposals. At inference we use only the semantic head; we train both heads jointly during fine-tuning, following SAM~3's training procedure~\cite{carion2025sam3segmentconcepts}:

Training jointly optimises the semantic and instance heads:
\begin{align}
  \mathcal{L}_{\text{sem}} &= 5\,\mathcal{L}_{\text{focal}}^{\alpha=0.6,\gamma=1.6} + 5\,\mathcal{L}_{\text{dice}} + r(e) \cdot 0.5\,\mathcal{L}_{\text{surf}} \nonumber \\
  \mathcal{L}_{\text{inst}} &= 5\,\mathcal{L}_{\text{focal}}^{\alpha=0.25,\gamma=2} + 5\,\mathcal{L}_{\text{dice}} + 2\,\mathcal{L}_{\text{cls}} + \mathcal{L}_{\text{pres}} \nonumber \\
  \mathcal{L} &= \mathcal{L}_{\text{sem}} + \mathcal{L}_{\text{inst}} \label{eq:sam3-loss}
\end{align}
Here $\mathcal{L}_{\text{focal}}$ is the focal loss with class-balancing factor $\alpha$ and focusing parameter $\gamma$, $\mathcal{L}_{\text{dice}}$ is the Dice loss, $\mathcal{L}_{\text{cls}}$ and $\mathcal{L}_{\text{pres}}$ are SAM~3's instance-head classification and presence losses, and $\mathcal{L}_{\text{surf}}$ The semantic-head loss adds $r(e) \cdot 0.5\,\mathcal{L}_{\text{surf}}$, with $r(e) = \min(1, e/15)$ ramping the surface loss linearly over the first 15 epochs. The surface term is $\mathbb{E}[\sigma(\hat{m}) \cdot d_{\text{norm}}]$, where $\sigma(\hat{m})$ is the predicted probability at each pixel (sigmoid of the model's per-pixel logit) and $d_{\text{norm}}$ is the unit-normalised signed distance map of the reference mask, negative inside the boundary and positive outside.

\paragraph{$k$-fold routing.}
The 208-case annotated pool is partitioned into five folds of roughly the same size. At inference, the SAM~3 loader activates the LoRA adapter from the fold that excluded the current case. This guarantees zero leakage of reference masks when reporting on cases that were in the LoRA training pool, which lets us benchmark on the full 208-case evaluation set.

\begin{table}[h]
\centering
\footnotesize
\setlength{\tabcolsep}{4pt}
\renewcommand{\arraystretch}{1.05}
\caption{SAM~3 fine-tuning configuration. Training is stopped by early-stopping on semantic-head validation IoU; the 20-epoch budget is never reached in practice.}
\label{tab:sam3-hparams}
\begin{tabular}{@{}l p{0.58\columnwidth}@{}}
\toprule
\textbf{Setting} & \textbf{Value} \\
\midrule
Backbone           & SAM~3 (frozen) \\
LoRA rank          & 16 \\
LoRA $\alpha$      & 32 (twice the rank) \\
LoRA dropout       & 0.05 \\
LoRA bias          & none \\
LoRA targets       & $\{q, k, v, o\}$ attention + $\{\text{fc}_1, \text{fc}_2\}$ MLP, on every transformer subsystem \\
Fully fine-tuned   & mask embedder, presence head, semantic projection \\
Training query     & ``planning boundary'' \\
Optimiser          & AdamW, lr $2 \times 10^{-4}$, weight decay $0.01$ \\
LR schedule        & Cosine decay to 5\% of base \\
Batch size         & 1, gradient accumulation 4 (effective 4) \\
Gradient clipping  & norm 0.1 \\
Precision          & bf16 \\
Epochs             & 20 max, patience 6 \\
\bottomrule
\end{tabular}
\end{table}

\subsection{Style-Transfer Augmentation}
\label{subsection:style-transfer}

\begin{figure}[h]
\centering
\includegraphics[width=\linewidth]{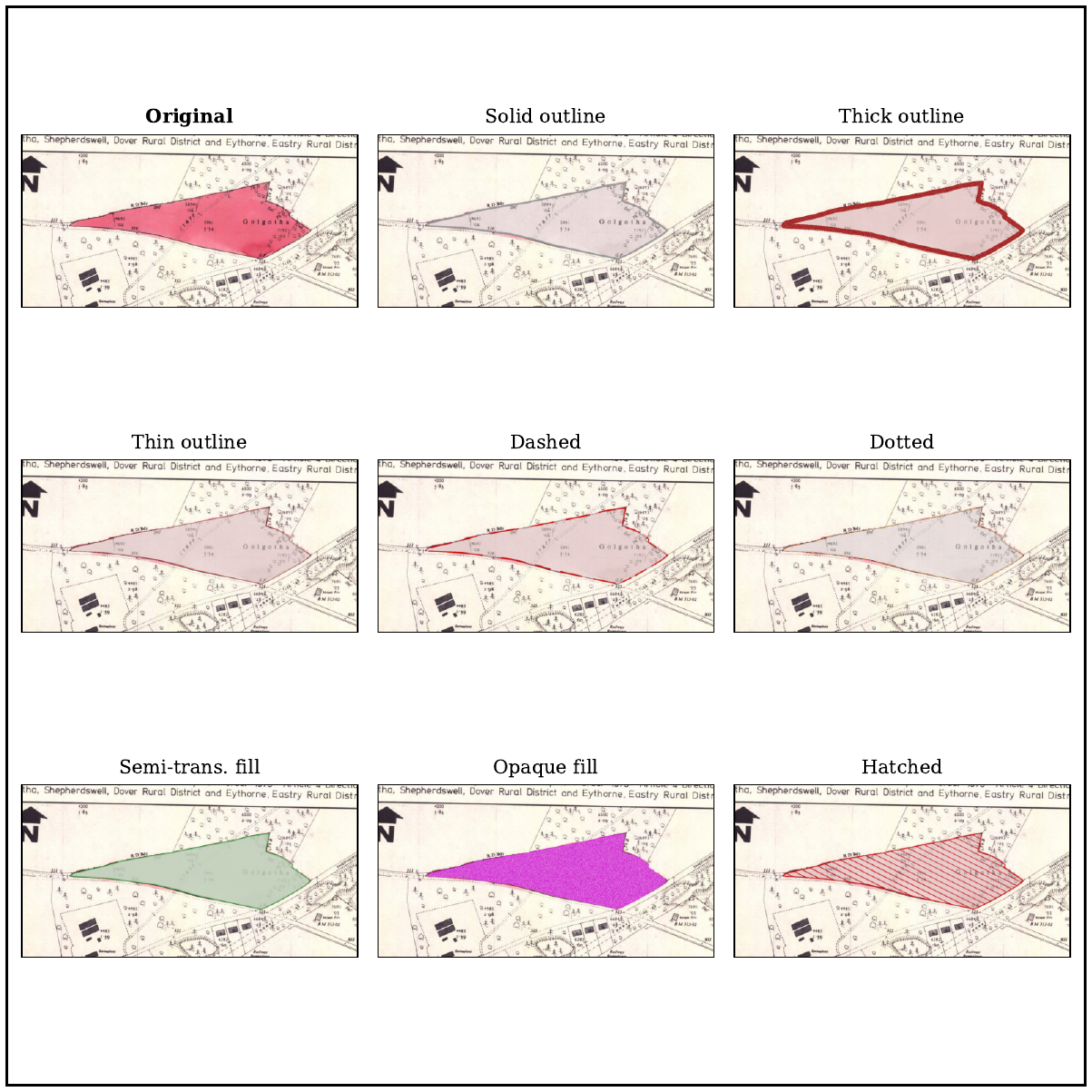}
\caption{Style-transfer augmentation applied to one training sample. Each panel except the top-left shows one of the eight pre-defined boundary styles drawn over the same map after the existing boundary has been faded into the background; colours and parameters are sampled randomly per panel from the ranges described in the text.}
\label{fig:style-transfer-aug}
\end{figure}

To widen effective training diversity from the small annotated pool, every training sample is re-rendered (with 50\% probability) in one of eight randomly-chosen boundary styles: solid outline at varying thicknesses (1--20\,px), dashed, dotted, semi-transparent fill ($\alpha \in [0.2, 0.5]$), fully-opaque fill ($\alpha = 1$), or diagonal hatching; \Cref{fig:style-transfer-aug} shows each style applied to one training sample. Eleven colours are sampled uniformly across red, blue, green, magenta, orange, black, and grey families. The existing boundary is first faded (desaturate 70--100\% and blend toward the local background mean), then the new style is drawn, and finally the result is roughened with Gaussian blur, additive noise, and stochastic morphological erosion or dilation to mimic scan artefacts. A random horizontal flip ($p=0.5$) and per-image brightness and contrast jitter ($p=0.5$, with each factor drawn uniformly at random from $[0.8, 1.2]$) are applied independently of the style transfer.

\subsection{SAM3-LoRA Per-Fold Cross-Validation}
\label{appendix:abl-segmentation-details}

\Cref{tab:sam3-cv} reports per-fold SAM3-LoRA validation IoU on the 208 cases within the dataset. The cross-fold mean reaches $0.912$ pixel IoU with low variance ($\sigma_{\mathrm{IoU}} = 0.028$).

\begin{table}[h]
\centering
\small
\begin{tabular}{@{}lccc@{}}
\toprule
\textbf{Fold} & \textbf{$|V|$} & \textbf{IoU} & \textbf{F1} \\ \midrule
0    & 43 & 0.911 & 0.943 \\
1    & 40 & 0.931 & 0.960 \\
2    & 42 & 0.879 & 0.909 \\
3    & 41 & 0.885 & 0.919 \\
4    & 42 & 0.952 & 0.973 \\ \midrule
Mean & 208 & \textbf{0.912} & \textbf{0.941} \\
Std  &     & 0.028 & 0.024 \\ \bottomrule
\end{tabular}
\caption{SAM3-LoRA 5-fold out-of-fold validation (semantic head). $|V|$ is the held-out validation size per fold.}
\label{tab:sam3-cv}
\end{table}

\subsection{Vanilla SAM~3 Prompt Search}
\label{subsection:abl-vanilla-sam-prompt-search}

Base SAM~3 is a promptable segmentation model; without LoRA fine-tuning, its output depends strongly on which English phrase is supplied as the query. We evaluated five candidate prompts on the full 208-case annotated pool: ``planning boundary'' (the LoRA-trained anchor), ``article 4 site boundary'', ``highlighted marked area'', ``site boundary'', and ``application site''. Across the five, ``highlighted marked area'' achieves the highest mean IoU (0.611), and is used as the vanilla SAM~3 baseline in \Cref{fig:abl-finetune}; ``planning boundary'' --- the LoRA-trained anchor --- is competitive on the distribution tail, with the highest median (0.886) and the highest fraction of cases at IoU $\geq 0.8$ (52.6\%).

\begin{table}[h]
\centering
\setlength{\tabcolsep}{1.5pt}
\small
\begin{tabular}{@{}lcccc@{}}
\toprule
\textbf{Prompt} & \textbf{Mean} & \textbf{Median} & \textbf{$\geq 0.5$} & \textbf{$\geq 0.8$} \\ \midrule
``highlighted marked area''  & \textbf{0.611} & 0.808 & 64.0\% & 51.2\% \\
``planning boundary''        & 0.582 & \textbf{0.886} & 60.2\% & \textbf{52.6\%} \\
``site boundary''            & 0.492 & 0.525 & 50.2\% & 44.1\% \\
``article 4 site boundary''  & 0.222 & 0.080 & 16.6\% & 14.7\% \\
``application site''         & 0.108 & 0.082 &  0.5\% &  0.0\% \\ \bottomrule
\end{tabular}
\caption{Vanilla SAM~3 prompt search on the 208 cases of the dataset.}
\label{tab:abl-sam-prompt-search}
\end{table}

\section{System Prompts}
\label{appendix:prompts}

Each LLM agent has a fixed system prompt and a Pydantic schema for its output. \Cref{fig:reader-prompt,fig:worker-prompt,fig:locate-prompt,fig:critic-prompt} reproduce the production prompts; \Cref{fig:vlm-direct-prompt} reproduces the single-shot prompt used by the VLM end-to-end baseline.

\begin{figure*}[t]
\begin{Verbatim}[frame=single,fontsize=\fontsize{7.5pt}{9pt}\selectfont,breaklines=true,breakanywhere=true,xleftmargin=0pt,xrightmargin=0pt]
You are a UK planning document reader. Read every page of the PDF carefully and populate the PDFInfo schema.

FIELD GUIDANCE (field descriptions in the schema are authoritative; these are additional rules):

- map_page_details: ONE MapPageMeta entry for EVERY page that contains any map-like or potentially-map content
  (both pages we want to position AND pages we discard).

  category: 'match' if this is a real positionable map with a drawn planning boundary on a cartographic
            background (OS-style, aerial, hand-drawn over OS). 'discard' otherwise.

  DISCARD AGGRESSIVELY -- false positives at the discard stage are worse than false negatives. A page is
  'discard' if ANY of: it is mostly text/forms/tables; it is a legend or key; it is a regional or town
  overview with no drawn boundary; it is a bare location pin or single-arrow inset; it is an indicative
  diagram without scale or cartographic detail; it is a photo or decorative illustration; it is a map
  background with no drawn planning boundary.

  area_group: -1 for discards. For 'match' pages, group pages that show the SAME geographic area under the
              same integer (0, 1, 2, ...). Different area_groups = different geographic areas; downstream
              projects each separately and UNIONS the resulting polygons.

  SCHEDULE-CLASSIFICATION GUARD: pages of the SAME geographic area shown for DIFFERENT permitted-development
  class restrictions of an Article 4 direction (Schedule 2 classes like Class A / Class E / Class F, Parts 1
  / 2, etc.) belong to the SAME area_group. The underlying physical area is the same; only the legal
  classification differs.

  boundary_clarity: 'clear' requires BOTH (a) the boundary line/hatch/edge is unambiguous to trace AND (b)
                    cartographic detail (streets, labels) is visible within and around the boundary.
                    Otherwise 'ambiguous'. 'none' = no boundary drawn.

  detail_level: close (parcel) / medium (neighbourhood) / wide (town or regional).
  caption:      one-line description (<=120 chars).

- map_pages: ranked list of category='match' page numbers. WITHIN AN AREA_GROUP -- primary picker (strict):
  Rule 1: higher boundary_clarity wins (clear > ambiguous > none). Rule 2 (on ties): prefer wider
  detail_level. Rule 3 (further tie): more cartographic detail. Ordering across different area_groups is
  arbitrary -- they will all be projected and unioned. Maps are usually near the end of the document.

- postcodes: extract ALL UK postcodes from site address, map title blocks, form fields, tables, and
  application metadata. Postcodes are the strongest geocoding signal.

- grid_refs: OS grid references on map edges (e.g. "TG 210 080", "TR 34 SE").

- is_district_wide: true if the planning boundary covers an entire administrative area (borough, district,
  ward, parish). Triggers include "Borough Wide Direction", "District Wide", "throughout the District of
  [name]". False for specific-site applications.

- district_name: if is_district_wide, the standard UK administrative name with a "UK" suffix. Examples:
  "Camden, UK", "Royal Borough of Kensington and Chelsea, UK". Provide "|"-separated alternates if ambiguous
  (e.g. "City of Westminster, UK | Westminster, UK").

- site_address: the SITE address (location of the boundary). IGNORE council, agent, or architect office
  addresses.

LOCATE-STAGE FIELDS (critical -- downstream geocoding relies on these):

- house_number_road_pairs: parse ANY house numbers + named roads. Collapse ranges -- "126, 128, 130, 132 and
  134 Norwich Road" -> ["126-134 Norwich Road"]. Skip OS parcel numbers.

- parish_names: bare strings. "in the parishes of Caistor St. Edmund and Keswick" -> ["Caistor St. Edmund",
  "Keswick"].

- admin_region: most specific admin unit, bare name. "in the District of South Norfolk" -> "South Norfolk".

- likely_town_or_city: best single answer for the town/city, synthesised from text, map labels, postcodes,
  ALL signals. Crucial for disambiguating common road names.

- visible_map_labels: what labels can you READ on the map image? Road names shown on roads, named buildings
  ("Colney Hall"), adjacent labelled places. Copy verbatim.

- adjacency_hints: named features touching/bordering the boundary from phrases like "adjoining X", "bordered
  by Y". Include only the named reference, not the preposition.
\end{Verbatim}
\caption{Reader system prompt.}
\label{fig:reader-prompt}
\end{figure*}

\begin{figure*}[t]
\begin{Verbatim}[frame=single,fontsize=\fontsize{7.5pt}{9pt}\selectfont,breaklines=true,breakanywhere=true,xleftmargin=0pt,xrightmargin=0pt]
You are the worker agent in a pipeline that extracts the application site boundary from UK planning permission
PDFs and projects it to a WGS84 GeoJSON polygon. The boundary is the area the applicant is requesting permission
for, marked on a site map within the PDF. Its visual style varies -- solid line, dashed, hatched, coloured fill.
A separate reader agent has already parsed the PDF and pre-rendered the map pages.

Your job: position the planning map against Ordnance Survey tiles using learned feature matching, then return
the projected polygon. SAM3 segmentation and GeoJSON projection are automatic -- you never call them explicitly.
Your tool surface is:
  propose_centers -> match_at(page=N, ...) -> commit_match -> BoundaryOutcome
plus lookup_district for documents covering an entire administrative area.

DOCUMENT STRUCTURE
- map_pages: ranked list of page numbers carrying a positionable map.
- area_group: every match page has an integer area_group. SAME group = duplicate views of the SAME geographic
  area; DIFFERENT groups = different geographic areas (multi-area documents).
- Each match_at call matches exactly ONE page (one area_group).
- Each commit_match commits exactly ONE candidate, i.e. one area_group. Multi-area documents loop
  propose_centers + match_at + commit_match SEPARATELY for each area_group; commits union into the running
  final result.
- Most documents are single-area -- run the loop once.

INPUT: PDFInfo summary + the top-ranked match page rendered upright. OUTPUT: a BoundaryOutcome with
status="accepted" or "district_lookup". Refusing a case is not supported; the pipeline always produces a polygon.

WORKFLOW

1. propose_centers() -- get one ranked candidate (lat/lon/sigma_m/source).

2. match_at(page=N, name, lat, lon). Returns candidate_id, area_group, page, n_inliers, scale_consistency,
   road_name_agreement (+ road_name_verdict), committed_groups, budget_remaining.

   THREE SIGNALS -- explicit tiers:

     n_inliers (RANSAC match strength):
       >= 100   STRONG     -- commit on this attempt unless another signal disagrees.
       50-99    OK         -- commit ONLY after trying at least one more candidate.
       25-49    WEAK       -- keep exploring; do not commit yet.
       < 25     TOO WEAK   -- try another candidate; never commit unless budget exhausted.

     scale_consistency (per-group, 0..1):
       >= 0.8   GOOD       -- recovered scale matches the reader's stated map scale.
       0.5-0.8  MARGINAL   -- scale stretched; prefer an alternative.
       < 0.5    BAD        -- trust only if n_inliers >= 100.

     road_name_agreement (per-group, 0..1):
       >= 0.6   STRONG     -- reader's road names found at this location.
       0.0      CONFLICT   -- OS has roads here but NONE match. Trust only if n_inliers >= 100.
       0.5      NEUTRAL    -- "no OS roads within radius"; sparse rural cartography.
       other    PARTIAL    -- some roads matched; weak corroboration.

   COMMIT DECISION (per area_group):
     1. STRONG n_inliers + GOOD scale + STRONG/NEUTRAL roads -> commit.
     2. Otherwise try another propose_centers candidate for this group BEFORE committing anything below STRONG.
     3. After 2+ attempts: highest n_inliers wins; tie-break on scale_consistency closer to 1.0, then on
        road_name_agreement.

3. commit_match(candidate_id) -- commits ONE candidate for its area_group. Re-calling overwrites the slot.

4. Return BoundaryOutcome with status="accepted" (or "district_lookup" if you took the lookup_district path).

BUDGET: max 5 match_at calls per case. NO INVENTED COORDINATES -- every (lat, lon) must come from
propose_centers. To add a missing place call propose_centers(extra_terms=["..."]). After a weak match, call
propose_centers(match_context="...") describing what went wrong; the locate sub-agent will pick from a
DIFFERENT signal type.
\end{Verbatim}
\caption{Worker system prompt.}
\label{fig:worker-prompt}
\end{figure*}

\begin{figure*}[t]
\begin{Verbatim}[frame=single,fontsize=\footnotesize,breaklines=true,breakanywhere=true,xleftmargin=0pt,xrightmargin=0pt]
You are the LOCATE STAGE for a UK planning permission boundary extraction pipeline.

Your job: given planning-document metadata (pdf_info text fields) AND the rendered planning map image,
produce ONE center coordinate (lat, lon) + an uncertainty radius sigma + confidence, so that downstream MINIMA
can refine it visually.

You have 1 offline geocoder tool:
- place(q, la=None) -- OS Open Names search (villages, schools, churches, named buildings).

PROTOCOL (every case):

1. VIEW the map image carefully. Look for labels, landmarks, distinctive features, road junctions, named
   buildings, hatched site polygon, neighbouring features. Note ANYTHING that's on the map but missing from
   pdf_info.

2. SCAN pdf_info. Priority of signals (most specific first):
   - house_number + named road in site_address
   - Named place / landmark from pdf_info OR from the map image
   - Parish name.

3. BUILD POOL via tool calls. Aim for 2-4 candidates from different signal types. Augment with terms FROM
   THE MAP IMAGE (do not limit yourself to pdf_info).

4. CLUSTER & PICK:
   - 2+ candidates within 500m -> tight consensus, sigma=200m, confidence='high'.
   - Single ambiguous (common place) -> sigma=800-1500m, confidence='med'.

5. EMIT the LocatePick to terminate. Once you have your pick, output the LocatePick directly as your final
   response -- do NOT make further tool calls. Be meticulous: copy the lat/lon EXACTLY from your strongest
   tool result -- do not paraphrase, do not round prematurely. Common bugs: (a) dropping a minus sign that
   should be there (-0.14 emitted as 0.14), (b) adding a minus sign that shouldn't be (+1.4 emitted as -1.4),
   (c) swapping top_lat and top_lon. Before emitting, verify the sign and order of the values against the
   tool result you are using.

BUDGET: <= 8 geocode tool calls per case. If you have made 8 calls, commit your best current guess with
confidence='low'.
\end{Verbatim}
\caption{Locate sub-agent system prompt.}
\label{fig:locate-prompt}
\end{figure*}

\begin{figure*}[t]
\begin{Verbatim}[frame=single,fontsize=\footnotesize,breaklines=true,breakanywhere=true,xleftmargin=0pt,xrightmargin=0pt]
You are an independent reviewer of a UK planning-boundary extraction pipeline. The agent has matched a
planning map to OS map tiles, generated candidate match attempts at different locations, projected SAM3
boundary masks through those matches, and committed one candidate as its final answer. Your job is pairwise
comparison across the stored candidates and a single directive on whether to accept or redirect.

WHAT YOU SEE
- ONE image per candidate, each LEFT|RIGHT:
    LEFT  = planning map with the SAM mask overlaid in translucent green. Label:
            "CANDIDATE {id} [COMMITTED] group {g} page {p}". The COMMITTED tag marks the worker's choice
            for THIS group.
    RIGHT = OS tile render at the matched window, projected polygon outlined in red.
            Label: "OS tile @ zoom={z}".
- Only the TOP-3 candidates by n_inliers are shown, plus the worker's committed candidate(s) if they fall
  outside the top-3.
- A metrics block accompanying the images, one line per candidate:
    "cand {id}  group {g}  page {p}  n_inliers={N}  road_name_agreement={r}  scale_consistency={s}  [COMMITTED]?"

WHAT "GOOD" LOOKS LIKE
Trace named roads, settlement shapes, or distinctive features between the planning map (left) and OS render
(right). The boundary line on the planning map should sit where the red outlined polygon sits in the OS
render. Road junctions, building blocks, water bodies should line up.

WHAT "BAD" LOOKS LIKE
- No road/feature correspondence: planning map shows urban streets; OS render shows farmland or a different
  street pattern.
- The SAM mask covers something other than the boundary (title block, legend, scale bar).
- The polygon outline lands well outside the planning map's drawn boundary, or its shape clearly does not match.

INTERPRETING THE METRICS (same tiers the worker uses):
  n_inliers:           >=100 STRONG / 50-99 OK / 25-49 WEAK / <25 TOO WEAK
  scale_consistency:   >=0.8 GOOD / 0.5-0.8 MARGINAL / <0.5 BAD (trust only if n_inliers >= 100)
  road_name_agreement: >=0.6 STRONG / 0.0 CONFLICT (trust only if n_inliers >= 100) / 0.5 NEUTRAL
                       (sparse cartography) / other PARTIAL
Numbers are supporting evidence -- the visual panels are the primary signal.

DECISION (pick exactly one action):
- approve       The worker's committed candidate(s) show clear correspondence AND the polygon outline aligns
                with the drawn boundary. Set chosen_candidate_id = committed_id.
- switch        A DIFFERENT stored candidate looks visually better FOR ITS area_group. Set chosen_candidate_id
                to that candidate; the worker's commit for that group will be replaced. Cite the visual
                feature that swayed you.
- retry_locate  None of the stored candidates show good correspondence -- they all appear to be in the wrong
                region. The worker will be asked to re-locate from a different geocoding signal.

OUTPUT
A single CriticDirective. Reasoning MUST cite a concrete observation -- a specific feature matched/mismatched,
or a specific metric. Never use vague language ('looks fine', 'reasonable').
\end{Verbatim}
\caption{Critic system prompt.}
\label{fig:critic-prompt}
\end{figure*}

\begin{figure*}[t]
\begin{Verbatim}[frame=single,fontsize=\footnotesize,breaklines=true,breakanywhere=true,xleftmargin=0pt,xrightmargin=0pt]
You are a UK planning permission boundary geocoder. Given a UK planning permission PDF, output a single
GeoJSON Feature whose geometry is a Polygon (single site) or MultiPolygon (multiple disjoint sites) covering
the APPLICATION SITE in WGS84 coordinates. NOT the council office that issued the document.

Think through these four steps before you write the output.

==================================================================================================
STEP 1 -- READ
==================================================================================================
Scan the PDF for every geographic signal you can find:
  * Site address (the location of the boundary), NOT the council/agent/architect office address.
  * UK postcodes inside the site address (format 'XX1 2YZ'). Ignore postcodes in council letterheads.
  * OS grid references (e.g. 'TG 210 080', 'TR 2648').
  * Named roads -- in the text or labelled on the map.
  * Named places -- parishes, villages, neighbourhoods, landmarks.
  * Labels printed on the map page itself.
  * Printed map scale (e.g. '1:2500').
  * Whether the boundary covers an entire borough / district / parish ('Borough Wide Direction',
    'throughout the District of X', etc.).

==================================================================================================
STEP 2 -- LOCATE
==================================================================================================
Convert your evidence into a single WGS84 anchor point for the site. UK longitudes range roughly -8.2 to 1.9;
UK latitudes range 49.8 to 60.9.

==================================================================================================
STEP 3 -- TRACE
==================================================================================================
Segment the drawn boundary on the planning map page. This is what you will project to WGS84 in STEP 4. Note:
  * Line style (red solid outline, hatched red, dashed blue, filled pink, black dot-dash).
  * Shape (rectangular, L-shaped, multiple disjoint parcels, elongated strip along the river, etc.).
  * If a printed scale is available, it can be useful for estimating the boundary's real-world size.

==================================================================================================
STEP 4 -- PROJECT
==================================================================================================
Translate the traced boundary into a WGS84 GeoJSON Feature anchored on the STEP 2 center, shaped and sized
per STEP 3.

OUTPUT FORMAT
  * type: 'Feature'
  * properties: free-form dict; may be empty.
  * geometry.type: 'Polygon' for a single site, 'MultiPolygon' for multi-area documents (Article 4
    directions, conservation areas covering multiple disjoint sites).
  * geometry.coordinates: a list of linear rings (Polygon) or a list of polygons each with their rings
    (MultiPolygon).

COORDINATE CONVENTION (do not get this wrong)
  * WGS84.
  * [longitude, latitude] order, NOT [latitude, longitude].
  * UK longitudes range -8.2 to 1.9; UK latitudes range 49.8 to 60.9.
  * Outer ring should close (first vertex == last); auto-closed if not.
  * Use 5 to 50 vertices per ring. Do NOT subdivide straight edges into many small segments -- a square
    needs 4 vertices, not 400.

Give your single best prediction. There is no follow-up. Be specific even when the document is ambiguous;
default to your most confident interpretation.
\end{Verbatim}
\caption{VLM end-to-end baseline prompt.}
\label{fig:vlm-direct-prompt}
\end{figure*}

\section{Declaration of AI use}
\label{appendix:ai_use}

We used AI assistants in this project for debugging the code, fixing grammar, and improving the writing of the paper. All AI-generated information was reviewed, verified, and edited by the authors, who take full responsibility for the content, correctness, and originality of the paper and the released artefacts.

\section{Artefact Use Consistent With Intended Use}
We use publicly available UK geospatial data (Code-Point Open, OS Open Names, OS BoundaryLine, OS OpenMap Local, OS Open Zoomstack) under Ordnance Survey's Open Government Licence (OGL v3), which explicitly permits use in research and derived products. The planning documents are sourced from public local authority records intended for public access. The Plan2Map dataset will be released for research purposes only, consistent with the access conditions of the underlying sources and with responsible data governance practices for AI systems \cite{kiden2024responsibleaigovernanceresponse}.

\end{document}